\documentclass[11pt]{article}

\usepackage[preprint]{acl}
\usepackage{float}
\usepackage{algorithm}
\usepackage{algpseudocode}
\usepackage{pgfplots}
\pgfplotsset{compat=1.18}
\usepackage{times}
\usepackage{latexsym}

\usepackage{tabularx}

\usepackage[T1]{fontenc}

\usepackage[utf8]{inputenc}

\usepackage{microtype}

\usepackage{inconsolata}

\usepackage{graphicx}
\usepackage{amsmath}
\usepackage{colortbl}
\usepackage{booktabs}
\usepackage{subcaption}
\usepackage{subcaption}
 \usepackage{amssymb}
\usepackage[font=tiny]{caption}
\title{Correct Diagnosis, Better Feedback: A Symbolic-Verifier for Faithful LLM Tutoring Feedback in Logic Proofs}

\author{
  Tahreem Yasir \\
  North Carolina State University \\
  \texttt{tyasir@ncsu.edu}
  \And
  Arnav Mody \\
  Purdue University \\
  \texttt{arnav2007mody@gmail.com}
  \AND
  Xiaoyi Tian \\
  Kennesaw State University \\
  \texttt{xtian@kennesaw.edu}
  \And
  Tiffany Barnes \\
  North Carolina State University \\
  \texttt{tmbarnes@ncsu.edu}
}

\begin{document}
\maketitle
\begin{abstract}
Effective LLM tutoring depends on correctly identifying the specific
error in a student's reasoning before generating feedback. We study
this problem in propositional-logic proof tutoring, where student
actions can be checked against formal inference rules. We introduce a
verifier-grounded architecture that separates diagnosis from language
generation. Using 600 balanced student actions, we compare a zero-shot
LLM detector, a fine-tuned detector, and a symbolic verifier. Each
diagnosis is processed by shared rationale and feedback agents,
isolating the effect of the initial diagnosis. The zero-shot detector
achieves a macro-F1 of 0.191; fine-tuning raises this to 0.709 but
retains systematic errors between structurally related classes.
Rationales generally preserve the diagnosis supplied to them, showing
that an incorrect diagnosis can be faithfully propagated through the
pipeline. Feedback can likewise remain faithful to its rationale,
non-revealing, and pedagogically appropriate while addressing the
wrong error. Verifier-grounded feedback achieves the highest diagnostic
correctness, and expert ratings largely uneven with the automatic
feedback evaluations. These findings show that apparent feedback
quality can conceal upstream diagnostic errors and that faithfulness
must be evaluated separately from correctness. Our code is publicly available.\footnote{\url{https://github.com/anonymouspaper64-debug/eacl_2026}}
\end{abstract}

\section{Introduction}
\label{sec:intro}
\begin{figure*}[t]
  \centering
  \includegraphics[width=2\textwidth, height=0.3\textheight, keepaspectratio]{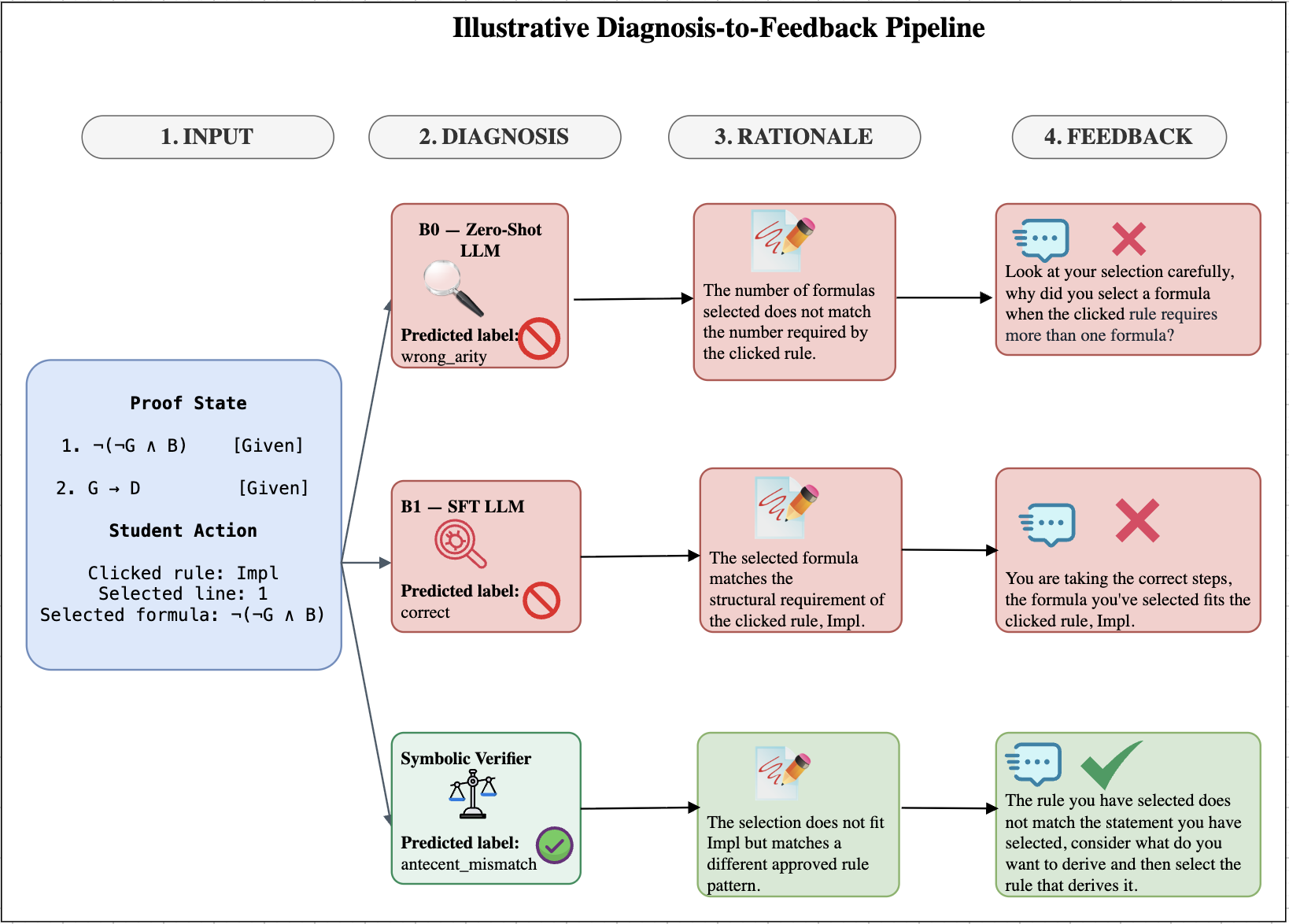}
  \caption{An illustrative example of one student action processed by
all three pipelines. Both LLM detectors misdiagnose the action: B0
predicts \texttt{wrong\_arity}, B1 predicts \texttt{correct}. The
symbolic verifier correctly identifies it as
\texttt{antecedent\_mismatch}. Each row shows the resulting rationale
and the feedback message shown to the student, tracing how an
incorrect diagnosis propagates into misleading feedback.}
  \label{fig:pipeline}
\end{figure*}

Effective tutoring feedback depends on correctly identifying students’ mistakes, and intelligent tutoring systems (ITS) support this process by diagnosing student reasoning at each problem-solving step, an intervention shown to improve learning \citep{vanlehn2011effectiveness}. However, conventional ITS depend on hand-crafted problems and feedback templates, which limits their scalability to new problem types \citep{zerkouk2025its_review}. Although LLMs reduce this authoring burden by generating feedback at scale, they remain unreliable at locating reasoning errors, thereby undermining the quality of the resulting feedback \citep{daheim2024stepwise, tyen2024llms, srivatsa2025llms, macina2023mathdial}. This limitation is particularly consequential in propositional logic, where the validity of each proof step is determined by a constrained set of inference rules. Prior studies of propositional-logic proof tutoring show that LLMs systematically reject correct reasoning and validate incorrect reasoning, producing inconsistent and hallucinated feedback \citep{yasir2026verification, yasir2026confirming}. These failures motivate separating error diagnosis from language generation and grounding the diagnosis in the formal structure of the proof.

Instead of relying on an LLM’s judgment, we propose a symbolic verifier that checks each student proof step against a formal specification of the inference rules and assigns the appropriate error label. Because a correct label does not yet constitute feedback, an LLM must translate the diagnosis into language the student can understand, potentially weakening, distorting, or contradicting it. Faithfulness therefore concerns whether the diagnosis is preserved in the generated feedback \citep{chuang2026faithlm}. Whereas prior work typically measures alignment between an explanation and a model’s own prediction or behavior, it cannot independently establish whether the explanation reflects a correct diagnosis \citep{jacovi2020towards}. By deriving labels through formal rule-based comparison, our verifier provides external ground truth for directly evaluating this preservation. To the best of our knowledge, this is the first approach to measure whether a correct diagnosis remains faithful when translated into feedback, rather than inferring faithfulness through multiple proxies.

To examine how diagnostic errors propagate into tutoring feedback, we
use a three-stage pipeline in which a detector assigns an error label,
a rationale agent explains the assigned label, and a feedback agent
translates the rationale into concise, Socratic guidance. We compare
three detectors: a zero-shot LLM, a fine-tuned LLM, and our symbolic
verifier. We evaluate rationale preservation using the its expressed support for
diagnostics label and feedback using two rubric-based LLM judges, whose
evaluations are compared with expert judgments. We address the
following research questions:

\noindent\textbf{RQ1.} How accurately do LLM-based error detectors
diagnose student proof errors, and where do they fail?

\noindent\textbf{RQ2.} Does the rationale faithfully preserve the
diagnosis provided by the detector?

\noindent\textbf{RQ3.} Does the resulting feedback faithfully preserve
the rationale?

\noindent\textbf{RQ4.} To what extent do the automatic feedback
evaluators agree with expert human judgments?

Our results show that both LLM-based detectors exhibit systematic
diagnostic failures. The rationale stage generally preserves the
diagnosis it receives, indicating that incorrect rationales primarily
originate from detection errors rather than rationale generation.
Although the feedback stage also largely preserves its input
rationale, feedback grounded in the symbolic verifier's labels
communicates the correct diagnosis more faithfully than feedback
produced using either LLM-based detector. 

\section{Related Work}
\subsection{LLM Error Diagnosis and Tutoring}
Several studies show that LLM tutors struggle to localize errors in student work \citep{daheim2024stepwise, tyen2024llms, srivatsa2025llms}. One explanation lies in their generative training objective, models are trained to continue plausible text and may therefore confirm reasoning that appears correct rather than detect reasoning that is flawed but plausible \citep{sonkar2024malalgoqa, yasir2026confirming}. Addressing this limitation in tutoring dialogues is further complicated because datasets with detailed pedagogical annotations remain scarce \citep{macina2023mathdial}. Supervised fine-tuning can help, but it does not close the gap. It can push models toward direct answers rather than scaffolded guidance \citep{sonkar2024pedagogical} and produce feedback that differs from teacher-written feedback in precision and specificity \citep{hsu2026mathedu}. Other approaches suggest to decompose the LLM tutor into separate components, isolating diagnosis from feedback generation and scaffolding \citep{kadir2026untamed}. Although this decomposition improves stage-level transparency, it does not establish whether each output is correct or whether the diagnosed error is preserved across stages. Our pipeline addresses this limitation by evaluating both the correctness of each stage and the preservation of the diagnosis across stages.

\subsection{Faithfulness of Explanation}
Faithfulness concerns whether an explanation accurately reflects the mechanism behind a prediction and is distinct from plausibility, which concerns whether an explanation merely appears convincing \citep{jacovi2020towards}. Natural-language and chain-of-thought explanations can appear convincing while failing to reflect what actually drove the prediction \citep{atanasova2023faithfulness, turpin2023language, zhao2025necessary}. Because faithfulness cannot be determined from the explanation alone, researchers have proposed indirect measures. These include perturbing inputs to test whether an explanation's stated reasons still hold \citep{chuang2026faithlm}, comparing causal and concept-based explanations of the same prediction \citep{matton2025walk}, testing an explanation's self-consistency \citep{parcalabescu2024measuring}, and using attribution methods sensitive to counterfactual changes in predicted probability \citep{siegel2024probabilities}. These methods share a common limitation: they rely on a model's internal or self-referential consistency because independently verified ground truth is unavailable. In our setting, formally verified diagnostic labels provide the independent ground truth needed to evaluate faithfulness directly rather than through model self-consistency.

\subsection{Symbolic Verification and Verified Feedback}
Intelligent tutors traditionally generate reliable feedback from fixed, template-based backends. Deep Thought, for example, combines data-driven problem selection with hint generation and strategy modeling for deductive logic proofs \citep{mostafavi2017evolution, barnes2010automatic}, while LOGAX generates formal proof paths to provide hints as students construct Hilbert-style proofs \citep{lodder2021generation}. Although these systems produce reliable feedback, they are difficult to extend to new problems and cannot scaffold reasoning that diverges from the tutor's predefined solution path \citep{yasir2026verification}. More recent LLM-based tutors reduce this rigidity. CodeAid uses an LLM to scale feedback in introductory programming courses, but it has been deployed in only one course, and its authors limit their validity claims to that restricted problem set rather than open-ended domains \citep{kazimcodeaid}. LeanTutor similarly combines an LLM with symbolic proof checking for logic proofs, but its checker relies on hand-written solutions and remains limited to a small problem set \citep{patel2026leantutor}. No prior work evaluates whether a correct diagnosis survives translation into feedback; we extend symbolic verification to test precisely this preservation.

\section{Dataset and Symbolic Verification}

\subsection{Dataset Construction}
\label{sec:dataset}

We construct our dataset from interaction logs of a GUI-based
propositional-logic tutor deployed in an undergraduate Discrete
Mathematics course at a large U.S.\ university in Fall 2025
\citep{barnes2010automatic}. The tutor is illustrated in
Appendix~\ref{app:DT}. The logs cover 32 proof problems across five
practice levels of increasing difficulty. Each logged action contains
the selected statements, clicked rule, and corresponding proof state.

Prior work identifies four recurring patterns in students' applications
of inference rules: selecting statements correctly, selecting the wrong
number of statements, selecting statements that fit a different rule,
and selecting statements that fit no rule
\citep{eagle2014exploration}. We operationalize these patterns as the
mutually exclusive labels \texttt{correct}, \texttt{wrong\_arity},
\texttt{antecedent\_mismatch}, and \texttt{wrong\_rule}, respectively. Table~\ref{tab:example1} shows one action from each class.
Our symbolic verifier (Section~\ref{sec:verifier}) assigns these labels
by unifying each selection against the tutor's rule set
\citep{robinson1965machine}. Because each diagnosis is determined by the action’s logical structure, the verifier provides ground-truth labels without requiring independent annotation.
\begin{table}[H]
  \caption{One student action per diagnosis class; the selected statement is bold.}
  \centering
  \small
  \begin{tabularx}{\columnwidth}{>{\raggedright\arraybackslash}p{0.50\columnwidth}>{\raggedright\arraybackslash}p{0.10\columnwidth}>{\raggedright\arraybackslash}X}
    \hline
    \textbf{Proof state} & \textbf{Rule} & \textbf{Label} \\
    \hline
    1.\ $(\neg O \lor L)\to(M \land \neg N)$ \newline
    2.\ $\mathbf{\neg O}$ \newline
    3.\ $K \to N$ &
    Add &
    \texttt{correct} \\
    \hline
    1.\ $(\neg O \lor L)\to(M \land \neg N)$ \newline
    2.\ $\neg O$ \newline
    3.\ $\mathbf{K \to N}$ &
    MP &
    \texttt{wrong\_arity} \\
    \hline
    1.\ $\mathbf{\neg(G \land A)}$ \newline
    2.\ $B \to A$ &
    Equiv &
    \texttt{antecedent\_} \newline \texttt{mismatch} \\
    \hline
    1.\ $K \to M$ \newline
    2.\ $Z \to R$ \newline
    3.\ $\mathbf{\neg(K \to R)}$ &
    DeM &
    \texttt{wrong\_rule} \\
    \hline
  \end{tabularx}
  \label{tab:example1}
\end{table}

The resulting corpus contains 2,000 labeled instances across the four
classes. Proof states range from two-line derivations to extended
multi-step proofs (2--25 statements; median 7), and all 15 inference
rules are represented. We partition the corpus into deduplicated,
disjoint subsets for evaluation and supervised fine-tuning. The
\emph{experiment set} contains 150 instances per class (600 total) and
serves as the balanced, stratified set used throughout
Section~\ref{sec:results}. We evaluate B0 and B1 against the verifier's
reference labels and compare downstream behavior across all three
pipelines. The \emph{SFT set} is drawn from the remaining traces, with
800 instances (200 per class) for training B1, 300 instances
(75 per class) for validation, and 300 instances (75 per class)
reserved for final testing.

\subsection{Problem Setup}
Let $\mathcal{R}$ be a set of propositional inference rules. Each rule
$r\in\mathcal{R}$ expects a fixed number of input statements, called
\emph{antecedents}, that match a fixed schema $\mathrm{ant}(r)$; its
\emph{arity} is the required number,
$\alpha(r)=|\mathrm{ant}(r)|$. Each rule derives a new statement; the
full rule set is provided in Appendix~\ref{tab:inference-rules}. Let
$\mathcal{R}_{\mathrm{ant}}\subseteq\mathcal{R}$ denote the rules whose
applicability can be determined from their antecedents alone. Given a
proof state $\sigma$ and a student action $(r,F)$, where
$r\in\mathcal{R}$ is the clicked rule and $F\subseteq\sigma$ contains
the selected statements, the verifier assigns one of the four
diagnosis labels defined above.

\subsection{Symbolic Verifier Construction}
\label{sec:verifier}
Given the antecedent schema $\mathrm{ant}(r)$, a rule applies when the
selected statements $F$ match the structure required by the clicked
rule $r$. We determine this fit using \emph{unification}
\citep{robinson1965machine}, which tests whether consistently
substituting statements for a rule's schema variables makes its
antecedents identical to the selected statements. A
\emph{substitution} maps the variables in $\mathrm{ant}(r)$ to
statements. We write $\mathrm{unify}(F,r)=\top$ if such a substitution
makes $\mathrm{ant}(r)$ and $F$ structurally identical. A variable
occurring more than once must receive the same binding at every
occurrence; it cannot be bound to different statements to force a
match.

The selection $Z\rightarrow R,\,Z$ unifies with MP's schema
$p\rightarrow q,\,p$ under $\{p\mapsto Z,\,q\mapsto R\}$. The
selection $Z\rightarrow R,\,W$ does not unify because $p$ would have
to bind to both $Z$ and $W$. For rules whose antecedent order is
irrelevant, we test all orderings. Because
$\mathrm{unify}(F,r)$ is determined solely by the structures of $F$
and $r$, it requires no LLM judgment and yields a deterministic
reference diagnosis.

\subsection{Diagnostic Algorithm}
\label{alg:diagnose}
Algorithm~\ref{alg:diagnose-algo} assigns a diagnosis by checking each
condition in order. It first checks whether the number of selected
statements matches the clicked rule's arity. If not, it returns
\texttt{wrong\_arity}. If the number however is correct, it uses $\mathrm{unify}(F, r)$ to test whether the
selection unifies with the clicked rule $r$;
a match returns
\texttt{correct}, and no further
check is needed. Otherwise, the algorithm tests $\mathrm{unify}(F, R_{ant})$ for every rule whose 
applicability can be determined from their antecedents. If the
selection matches another rule, it returns
\texttt{antecedent\_mismatch}; however, if no rule matches, it returns
\texttt{wrong\_rule}.
 
The verifier checks only whether the selected statements fit the rule's
antecedents. It does not check the derived statement because the
tutor's GUI generates it automatically from the rule and selection.
To validate the labels, two annotators independently labeled a
stratified sample of 100 instances while blind to the verifier's
outputs. Inter-annotator agreement was $\kappa>.80$. Their agreement
with the verifier was 95\% and 92\%, respectively. Annotation details
are provided in Appendix~\ref{app:annotation}.
 
\begin{algorithm}[t]
\small
\caption{Student error diagnosis}
\label{alg:diagnose-algo}
\begin{algorithmic}[1]
\Statex \textbf{Input:} selection $F$, clicked rule $r$
\If{$|F| \neq \alpha(r)$}
  \State \Return \texttt{wrong\_arity}
    \Comment{wrong antecedent count}
\EndIf
\If{$\mathrm{unify}(F,r)$}
  \State \Return \texttt{correct}
    \Comment{$F$ fits the clicked rule}
\EndIf
\For{each $r'\in\mathcal{R}_{\mathrm{ant}}\setminus\{r\}$}
  \If{$\mathrm{unify}(F,r')$}
    \State \Return \texttt{antecedent\_mismatch}
      \Comment{fits another rule}
  \EndIf
\EndFor
\State \Return \texttt{wrong\_rule}
  \Comment{fits no rule}
\end{algorithmic}
\end{algorithm}
\section{Methods}
We compare three feedback pipelines that differ only in how the
student's error is detected, allowing us to isolate the effect of
diagnosis and assess whether it is preserved through downstream
generation. Each instance comprises a proof state $\sigma$ and a
student action $(r,F)$ (Section~\ref{sec:dataset}). Each pipeline has
three stages: an error detector assigns one of four diagnosis labels
(Section~\ref{sec:verifier}), a rationale agent explains the assigned
label, and a feedback agent converts the rationale into scaffolded,
student-facing guidance. All LLM-based components within our main pipeline use Llama 3.1 8B
Instruct. The rationale and feedback agents use identical
prompts, and settings across pipelines, leaving the detector
as the only experimental variable.

\subsection{Error Detection}
We compare three detectors: the base LLM, its task-specific fine-tuned
counterpart, and our symbolic verifier. Each processes the same
instance, comprising proof state $\sigma$ and student action $(r,F)$,
and returns one of the four diagnosis labels defined in
Section~\ref{sec:verifier}. This comparison evaluates zero-shot
diagnosis, the effect of supervised fine-tuning, and the effect of
verifier-grounded diagnosis on the resulting rationale and feedback.

- \noindent\textbf{Zero-Shot LLM (B0):}
Our first detector uses the base Llama 3.1 8B Instruct without
task-specific fine-tuning. Its prompt provides $\sigma$, $(r,F)$,
tutor-specific notation, inference-rule definitions, arity
requirements, and the four-class taxonomy. The model returns a single
diagnosis label. The prompt is provided in
Appendix~\ref{app:prompt_error_detection}.

- \noindent\textbf{Supervised Fine-Tuning (B1):} B1 fine-tunes the same base checkpoint with QLoRA
\citep{dettmers2023qlora} on 800 verifier-labeled training instances.
It uses the same inference prompt and output format as B0, making
supervised fine-tuning the only difference between the two LLM
detectors. We select the checkpoint with the lowest validation loss;
full hyperparameters, training schedule, and loss curves appear in
Appendix~\ref{app:sft_details}.

- \noindent\textbf{Verifier Grounded Pipeline:}
Our proposed pipeline replaces LLM-based detection with the symbolic
verifier described in Section~\ref{sec:verifier}. The verifier assigns
a diagnosis using Algorithm 1 describe in ~\ref{alg:diagnose} yielding a deterministic reference diagnosis. This
condition tests whether that diagnosis is preserved through rationale
and feedback generation.
 
\subsection{Rationale Agent}
The rationale agent receives student action
$(r,F)$, and detector-assigned diagnosis, which it treats as
authoritative. It is instructed to explain this diagnosis rather than
independently reclassify the action. Because free-form generation may
diverge from its input, we define a rationale as \emph{faithful} when
it preserves the assigned diagnosis, regardless of whether that
diagnosis matches the verifier's reference diagnosis. We evaluate
diagnostic correctness separately by comparing the diagnosis expressed
in the rationale with the reference diagnosis
(Section~\ref{sec:rationale-faithfulness}). The same prompt is used
across all pipelines and is provided in
Appendix~\ref{app:prompt_rationale_agent}.

\subsection{Feedback Agent}
The feedback agent receives student action
$(r,F)$, and rationale, but not the detector-assigned or
reference diagnosis. It is instructed to preserve the rationale's
diagnosis rather than independently reclassify the action. This
prevents direct label leakage while retaining the context needed to
produce concrete guidance. The output consists of two to three
sentences of Socratic guidance that may identify the relevant issue
but must not reveal specific details such as mentioning a corrective rule, provide the derived formula, or
reveal the next step. The same prompt is used across all pipelines and
is provided in Appendix~\ref{app:prompt_feedback_agent}.

\section{Evaluation}
We evaluate each pipeline stage on the 600-instance experiment set:
B0 and B1 for diagnostic correctness against the verifier's reference
diagnoses; rationales for preservation of their assigned diagnoses and
diagnostic correctness; and feedback for rationale preservation,
diagnostic correctness, non-revelation, and pedagogical
appropriateness. We then compare the automatic feedback evaluations
with expert judgments.
 
- \noindent\textbf{Diagnostic Correctness (RQ1): }
We compare the labels predicted by B0 and B1 with the verifier's
reference diagnoses. We report macro-F1 as the primary metric because
it gives equal weight to all four classes, along with per-class
precision, recall, and F1 to identify class-specific errors. The
verifier-grounded pipeline uses the reference diagnosis by construction
and is therefore not evaluated as an independent detector.

- \noindent\textbf{Faithfulness of the Rationale (RQ2): }
\label{sec:rationale-faithfulness}
Once a diagnosis is assigned, we measure whether the rationale
preserves it. String matching is insufficient because the rationale
expresses the diagnosis in prose. We therefore test whether its text
entails statements defining each diagnosis class. Natural language
inference (NLI) models estimate whether one text semantically supports
another. We use BART-large-MNLI~\citep{lewis-etal-2020-bart} as a
zero-shot classifier over the four diagnosis classes.
 
For each class \(c\), domain experts define a set \(T_c\) of five
declarative hypotheses covering its defining properties and common
linguistic formulations. The four sets correspond to the mutually
exclusive diagnosis classes; the complete hypotheses appear in
Appendix~\ref{app:nli-templates}. For each rationale \(x_i\), the
model produces five entailment scores per class. We retain the
highest-scoring template for each class and normalize the resulting
four scores using a softmax:
\vspace{-6mm}
\begin{equation}
\small
z_{ic} = \max_{t\in T_c}\ell_{ict}.
\label{eq:max-template-logit}
\end{equation}
\vspace{-3mm}
\begin{equation}
\small
s_{ic} = \frac{\exp(z_{ic})}{\sum_{c'=1}^{4}\exp(z_{ic'})},
\qquad
\widetilde{c}_i = \operatorname*{arg\,max}_{c} s_{ic}.
\label{eq:rationale-classification}
\end{equation}
%
Here \(\ell_{ict}\) is the entailment logit for rationale \(i\) against
template \(t\) of class \(c\), \(z_{ic}\) is the strongest such logit,
\(s_{ic}\) is the normalized support for class \(c\), and
\(\widetilde{c}_i\) is the NLI-assigned diagnosis. Let
\(\widehat{c}_i\) denote the detector-assigned diagnosis and \(c_i^*\)
the verifier's reference diagnosis. We define
\begin{equation}
\small
F_{\mathrm{sys}} = \frac{1}{N}\sum_{i=1}^{N}\mathbb{I}[\widetilde{c}_i=\widehat{c}_i],
\quad
F_{\mathrm{gt}} = \frac{1}{N}\sum_{i=1}^{N}\mathbb{I}[\widetilde{c}_i=c_i^*].
\label{eq:rationale-faithfulness}
\end{equation}
\(F_{\mathrm{sys}}\) measures whether the rationale is faithful to its
assigned diagnosis, whereas \(F_{\mathrm{gt}}\) measures whether it
expresses the verifier's reference diagnosis. We additionally report mean
normalized support for both diagnoses:
\begin{equation}
\small
\overline{s}_{\mathrm{sys}} = \frac{1}{N}\sum_{i=1}^{N}s_{i,\widehat{c}_i},
\qquad
\overline{s}_{\mathrm{gt}} = \frac{1}{N}\sum_{i=1}^{N}s_{i,c_i^*}.
\label{eq:mean-nli-support}
\end{equation}
We treat these scores as automated indicators of rationale--diagnosis
agreement, not measures of formal logical validity. As a manual check
of diagnosis preservation, one annotator reviewed a sample of 200
rationales stratified by assigned diagnosis (50 per class) and judged
whether each clearly conveyed the diagnosis it was generated to
explain. We report the results in Section~\ref{sec:results}.

\noindent\textbf{Feedback Evaluation (RQ3).}
\phantomsection
\label{sec:feedback-faithfulness}
Unlike rationales, feedback must both preserve the diagnosis and
satisfy pedagogical constraints that class-specific NLI templates do
not capture. We therefore use two rubric-based LLM judges,
implemented with GPT-5.1, to score each dimension from 0 to 2.
Table~\ref{tab:judge-rubric} defines the scoring criteria. We report
each dimension separately.

\begin{table}[H]
  \centering
  \tiny
  \caption{Scoring rubric for the two feedback judges. Reference
  diagnosis denotes the verifier's \(c_i^*\).}
  \begin{tabularx}{\columnwidth}{>{\raggedright\arraybackslash}p{0.15\columnwidth}XXX}
    \toprule
    \textbf{Dimension} & \textbf{2} & \textbf{1} & \textbf{0} \\
    \midrule
    \multicolumn{4}{l}{\textit{\textbf{Quality judge}
    (feedback vs.\ corresponding rationale)}} \\
    \midrule
    \textbf{Rationale
    faithfulness} &
      Preserves the rationale's diagnosis &
      Partial or vague &
      Contradicts or changes it \\
    \addlinespace
    \textbf{Non-revelation} &
      No corrective rule, formula, or next step &
      Reveals a substantial clue &
      Explicitly reveals one \\
    \addlinespace
    \textbf{Pedagogical
    appropriateness} &
      Concrete, relevant, appropriate depth &
      Generic or mismatched depth &
      Irrelevant or misleading \\
    \midrule
    \multicolumn{4}{l}{\textit{\textbf{Diagnostic judge}
    (feedback vs.\ reference diagnosis)}} \\
    \midrule
    \textbf{Diagnostic
    correctness} &
      Communicates the reference class's defining property &
      Compatible but underspecified &
      Contradicts it or targets another class \\
    \bottomrule
  \end{tabularx}
  \label{tab:judge-rubric}
\end{table}

The quality judge receives student action, rationale,
and feedback, but not the assigned or reference diagnosis. It is
instructed to assess whether the feedback preserves the rationale
rather than independently re-diagnosing the action, thereby measuring
translation quality rather than correctness. The diagnostic judge
receives the student action, feedback, and reference
diagnosis, but not the rationale or assigned diagnosis. It also
receives the verifier's fixed taxonomy so that it can assess the accuracy of assigned label.

Together, the judges distinguish errors inherited from the rationale
from errors introduced during feedback generation. Their
prompts are provided in
Appendix~\ref{app:prompt_feedback_quality_judge} and
Appendix~\ref{app:prompt_diagnostic_accuracy_judge}.

\noindent\textbf{Human Validation (RQ4).}
To validate the automatic judges, two annotators, blind to their
outputs, first completed a joint training and calibration phase,
independently scoring the same feedback instances with the judge
rubric until reaching 89\% inter-rater agreement. The first annotator
then scored the remaining stratified sample. We compare these human
ratings with the corresponding automatic judge scores in
Section~\ref{sec:results}.
\section{Results}
\label{sec:results}

We organize the results around detector accuracy and error patterns
(RQ1), rationale faithfulness (RQ2), feedback faithfulness and
correctness (RQ3), and agreement with human judgments (RQ4).

\subsection{RQ1: Diagnostic Accuracy and Error Patterns}
\label{sec:rq1}
Table~\ref{tab:perclass} reports macro-F1 and per-class precision,
recall, and F1 for B0 and B1 on the 600-instance experiment set. B0
achieves a macro-F1 of 0.191, below a 0.25 uniform-random reference,
whereas B1 reaches 0.709, an absolute improvement of 0.518.

\begin{table}[H]
  \centering
  \caption{Per-class precision, recall, and F1 for B0 and B1 on the
  600-instance experiment set (150 per class).}
  \label{tab:perclass}
  \tiny
  \begin{tabular}{lcccccc}
    \hline
    & \multicolumn{3}{c}{\textbf{Zero-shot (B0)}} & \multicolumn{3}{c}{\textbf{SFT (B1)}} \\
    
    \hline
    \textbf{Macro-F1} & \multicolumn{3}{c}{\textbf{0.191}} & \multicolumn{3}{c}{\textbf{0.709}} \\
    \hline
    \textbf{Class} & \textbf{P} & \textbf{R} & \textbf{F1} & \textbf{P} & \textbf{R} & \textbf{F1} \\
    \textbf{correct}              & 0.12 & 0.01 & 0.01 & 0.75 & 0.73 & 0.74 \\
    \textbf{wrong\_arity}          & 0.29 & 0.19 & 0.23 & 0.98 & 0.82 & 0.89 \\
    \textbf{antecedent\_mismatch}  & 0.25 & 0.43 & 0.31 & 0.81 & 0.37 & 0.51 \\
    \textbf{wrong\_rule}           & 0.17 & 0.27 & 0.21 & 0.74 & 0.65 & 0.69 \\
    \hline
  \end{tabular}
\end{table}

B0's poor performance is most evident on \texttt{correct}, for which
recall is only 0.01. B1 raises this recall to 0.73 and improves every
class, performing best on \texttt{wrong\_arity} (F1=0.89). Its main
remaining weakness is \texttt{antecedent\_mismatch} (F1=0.51):
B0 predicts this class broadly (R=0.43, P=0.25), whereas B1 predicts
it conservatively (R=0.37, P=0.81). The verifier is not evaluated as
an independent detector because its outputs define the reference
diagnoses and therefore have precision, recall, and F1 of 1.0 by
construction.

\subsection{RQ2: Faithfulness of the Rationale}
\label{sec:rq2}

Table~\ref{tab:entailment_overall} reports rationale faithfulness to
the assigned diagnosis (\(F_{\mathrm{sys}}\)), correctness against the
reference diagnosis (\(F_{\mathrm{gt}}\)), and mean normalized support. $F_{\mathrm{sys}}$ is similar for B0 and B1
(73.5\% and 75.8\%) but substantially higher for the
verifier-grounded pipeline (94.3\%). Thus, rationales often preserve
their assigned diagnosis, even when that diagnosis is incorrect.
$F_{\mathrm{gt}}$ follows the detector ranking in RQ1, decreasing from
94.3\% for the verifier to 60.3\% for B1 and 25.3\% for B0. Because
the verifier's assigned and reference diagnoses coincide by
construction, its 5.7-point drop arises during rationale generation.
Normalized support shows the same pattern: the scores coincide for the
verifier (0.977), whereas B1 (0.660 vs.\ 0.566) and B0
(0.573 vs.\ 0.256) support the assigned diagnosis more strongly than
the reference diagnosis.

\begin{table}[H]
  \centering
  {\small
  \caption{Rationale faithfulness by pipeline. For the verifier, label accuracy is 100\%, hence
  $F_{\mathrm{sys}}=F_{\mathrm{gt}}$; the drop to 94.3\% reflects label to rationale
  rendering loss.}
  \label{tab:entailment_overall}}
  \small
  \begin{tabular}{lcccc}
    \hline
    \textbf{System}
    & $F_{\mathrm{sys}}$
    & $F_{\mathrm{gt}}$
    & $\overline{s}_{\mathrm{sys}}$
    & $\overline{s}_{\mathrm{gt}}$ \\
    \hline
    Verifier (proposed)
      & 94.3\% & 94.3\% & 0.977 & 0.977 \\
    B1 (SFT)
      & 75.8\% & 60.3\% & 0.660 & 0.566 \\
    B0 (zero-shot)
      & 73.5\% & 25.3\% & 0.573 & 0.256 \\
    \hline
  \end{tabular}
\end{table}

\begin{figure*}[t]
\centering
\setlength{\tabcolsep}{4pt}

\subcaptionbox{
  B0 (zero-shot),
  $F_{\mathrm{gt}}=25.3\%$
  \label{fig:cm-b0}
}[0.32\linewidth]{
\footnotesize
\begin{tabular}{l|cccc}
\toprule
& \textbf{WA} & \textbf{AM}
& \textbf{WR} & \textbf{C} \\
\midrule
\textbf{WA}
& \cellcolor{blue!45}62
& \cellcolor{orange!30}74
& \cellcolor{orange!30}14
& 0 \\

\textbf{AM}
& \cellcolor{orange!30}51
& \cellcolor{blue!45}84
& \cellcolor{orange!30}15
& 0 \\

\textbf{WR}
& \cellcolor{orange!30}34
& \cellcolor{orange!30}111
& \cellcolor{blue!45}5
& 0 \\

\textbf{C}
& \cellcolor{orange!30}83
& \cellcolor{orange!30}58
& \cellcolor{orange!30}8
& \cellcolor{blue!45}1 \\
\bottomrule
\end{tabular}
}
\hfill
\subcaptionbox{
  B1 (SFT),
  $F_{\mathrm{gt}}=60.3\%$
  \label{fig:cm-sft}
}[0.32\linewidth]{
\footnotesize
\begin{tabular}{l|cccc}
\toprule
& \textbf{WA} & \textbf{AM}
& \textbf{WR} & \textbf{C} \\
\midrule
\textbf{WA}
& \cellcolor{blue!45}150
& 0
& 0
& 0 \\

\textbf{AM}
& \cellcolor{orange!30}27
& \cellcolor{blue!45}89
& \cellcolor{orange!30}25
& \cellcolor{orange!30}9 \\

\textbf{WR}
& \cellcolor{orange!30}96
& \cellcolor{orange!30}13
& \cellcolor{blue!45}39
& \cellcolor{orange!30}2 \\

\textbf{C}
& \cellcolor{orange!30}18
& \cellcolor{orange!30}41
& \cellcolor{orange!30}7
& \cellcolor{blue!45}84 \\
\bottomrule
\end{tabular}
}
\hfill
\subcaptionbox{
  Verifier (ground truth),
  $F_{\mathrm{gt}}=94.3\%$
  \label{fig:cm-verifier}
}[0.32\linewidth]{
\footnotesize
\begin{tabular}{l|cccc}
\toprule
& \textbf{WA} & \textbf{AM}
& \textbf{WR} & \textbf{C} \\
\midrule
\textbf{WA}
& \cellcolor{blue!45}150
& 0
& 0
& 0 \\

\textbf{AM}
& \cellcolor{orange!30}20
& \cellcolor{blue!45}128
& \cellcolor{orange!30}2
& 0 \\

\textbf{WR}
& \cellcolor{orange!30}1
& \cellcolor{orange!30}11
& \cellcolor{blue!45}138
& 0 \\

\textbf{C}
& 0
& 0
& 0
& \cellcolor{blue!45}150 \\
\bottomrule
\end{tabular}
}
\caption{Ground-truth confusion matrices for the rationale evaluation.
Rows are verifier ground-truth classes, columns are NLI-assigned
rationale classes. WA = \texttt{wrong\_arity}, AM =
\texttt{antecedent\_mismatch}, WR = \texttt{wrong\_rule}, C =
\texttt{correct}.}
\label{fig:confusion_gt}
\end{figure*}

Figure~\ref{fig:confusion_gt} reveals distinct class-level patterns.
B0 assigns only 1 of 150 \texttt{correct} instances to the correct
rationale class and maps 111 of 150 \texttt{wrong\_rule} instances to
\texttt{antecedent\_mismatch}. B1 improves substantially, although its
largest confusion maps 96 \texttt{wrong\_rule} instances to
\texttt{wrong\_arity}. As a manual check, an annotator judged 88\% of
the sampled rationales as clearly conveying their assigned diagnosis;
the remaining 12\% were unclear or ambiguous.

\subsection{RQ3: Feedback Faithfulness and Diagnostic Correctness}
\label{sec:rq3}

Table~\ref{tab:combined-quality} reports feedback faithfulness,
pedagogical appropriateness, and diagnostic correctness. B0 and B1 are
further partitioned by whether the detector-assigned diagnosis matches
the verifier's reference diagnosis. The verifier-grounded condition is
not partitioned because its assigned and reference diagnoses coincide
by construction.

\begin{table}[H]
  \centering
  \small
  \caption{Feedback faithfulness, pedagogical appropriateness, and
  diagnostic correctness on a 0--2 scale. Non-revelation is omitted
  because all rows score 2.000.}
  \label{tab:combined-quality}
  \begin{tabular}{llrccc}
    \toprule
    \textbf{System} & \textbf{Label} & \textbf{$N$}
    & \textbf{Faith.} & \textbf{Pedag.} & \textbf{Diag.} \\
    \midrule
    Verifier & --      & 600 & 1.710 & 1.292 & 1.775 \\
    \midrule
    B1 & Overall        & 600 & 1.625 & 1.263 & 1.513 \\
    B1 & Matched         & 386 & 1.672 & 1.332 & 1.734 \\
    B1 & Mismatched      & 214 & 1.472 & 1.042 & 0.803 \\
    \midrule
    B0 & Overall        & 600 & 1.617 & 1.040 & 1.060 \\
    B0 & Matched         & 135 & 1.672 & 1.061 & 1.794 \\
    B0 & Mismatched      & 465 & 1.601 & 1.034 & 0.855 \\
    \bottomrule
  \end{tabular}
\end{table}

Across the overall rows, the verifier-grounded pipeline ranks highest
on all three dimensions, followed by B1 and B0. We compare matched and mismatched instances within each LLM pipeline
using Mann--Whitney $U$ tests; full statistics appear in
Appendix~\ref{app:rq3-stats}. For B1, mismatched diagnoses receive
significantly lower scores on all three dimensions, with the largest
drop in diagnostic correctness (1.734 to 0.803). For B0, diagnostic
correctness similarly decreases from 1.794 to 0.855, whereas
faithfulness and pedagogical appropriateness change little and are not
significantly different. Thus, B0 can produce apparently faithful and pedagogically appropriate
feedback explaining an incorrect diagnosis, making the error
detectable only through comparison with the reference diagnosis.
 
\subsection{RQ4: Agreement with Human Judgments}
For human validation, agreement was strongest for
non-revelation and diagnostic correctness, followed by rationale
faithfulness. Agreement on pedagogical appropriateness was markedly
lower, suggesting that this dimension was more subjective and less
consistently interpreted. Overall, the diagnostic judge aligned more
closely with human judgments, whereas the quality judge was more useful
for assessing rationale preservation and constraint compliance than
overall feedback quality.

\section{Discussion}
\label{sec:discussion}
Our results show that diagnostic correctness and feedback quality are
distinct properties. Fine-tuning substantially improves detection but
does not recover the verifier's explicit comparison between the
clicked rule and alternative applicable rules. Consistent with prior
work, LLMs struggle more to locate an error than to explain one once
identified \citep{daheim2024stepwise, tyen2024llms}. Diagnosis is
therefore the main source of error, although rationale and feedback
generation are not lossless.

\textbf{Architectural implications}
Our findings suggest a clear division of labor for neuro-symbolic
tutoring systems. When an error type is locally decidable, a symbolic
component should determine what is wrong, while an LLM should decide
how to communicate it. Replacing the verifier with a fine-tuned
classifier improves flexibility but gives up the structural guarantees
available from explicit rule checking. The verifier's role
is therefore to ensure that language generation begins from a
formally grounded diagnosis.

\textbf{The rationale as an audit boundary}
The explicit rationale is important even though it is not shown to the
student. It creates an inspectable boundary between diagnosis and
feedback generation. This makes it possible to locate the translation quality during rationale to feedback generation. A pipeline that generates feedback
directly from the student action would hide these distinctions behind
a single fluent response. Future tutoring systems should preserve such
an intermediate representation whenever correctness must be traced
across multiple components.

\textbf{Tracing Correctness Across Stages}
Prior work often includes correctness alongside pedagogical quality in
human feedback evaluation. Our results show that end-to-end correctness
alone does not reveal where an error originates. In the zero-shot
pipeline, feedback can remain faithful to its rationale and
pedagogically appropriate while preserving an incorrect upstream
diagnosis. Evaluating diagnostic correctness, rationale faithfulness,
and pedagogical quality separately therefore makes error propagation
across the pipeline visible. Human evaluation supports treating the two judges as complementary
rather than interchangeable. The quality judge assesses whether
feedback preserves its rationale and satisfies generation constraints,
but these properties do not establish diagnostic correctness. The
diagnostic judge showed stronger alignment with human ratings, while
the low agreement on pedagogical appropriateness highlights the
subjectivity of evaluating instructional quality.

\textbf{Future Directions}
More broadly, our proposed method needs only a local decision point where
correctness can be checked by a sound external procedure. We hypothesize that this design where symbolic
verification for diagnosis and LLMs for language generation could provide a lightweight
and scalable route beyond propositional logic. For example checking algebraic simplification by computing variables' values, programming assignments grading against test
cases, or database queries evaluation against expected results. In such settings, symbolic verification could serve both
as a system component and as a measurement instrument for locating
where correctness is lost.


\section{Conclusion}
Tutoring feedback is useful only when it addresses the student's actual
error. Using student interaction logs from a propositional-logic tutor,
we evaluated a three-stage pipeline separating error detection,
rationale generation, and feedback generation. We show that
fine-tuning substantially improves LLM-based diagnosis but retains
systematic errors between structurally related classes. Replacing the
LLM detector with a symbolic verifier grounds generation in the formal
proof structure. Our results further show that faithful, non-revealing,
and pedagogically appropriate feedback can still preserve an incorrect
diagnosis. The verifier's primary benefit is therefore not better
language generation, but ensuring that plausible feedback addresses
the student's actual error. More broadly, our verifier-grounded evaluation separates two questions:
whether the rationale and feedback preserve the diagnosis assigned by
the system, and whether that diagnosis agrees with the verifier's
independently established reference label. This
distinction can support the evaluation of feedback pipelines in other
domains where step-level correctness or error type is locally decidable
through a sound external checker.
\section*{Limitations}

Our study examines diagnosis-to-feedback propagation in a deliberately
constrained setting. The data come from one propositional-logic tutor,
one undergraduate course, and a fixed set of proof problems and
inference rules. The balanced experiment set also differs from the
natural distribution of student errors. Moreover, the four-class
taxonomy captures local rule-application errors but not broader
misconceptions, proof-planning failures, or errors in derived
conclusions.
The findings therefore apply most directly to settings in which
correctness is locally decidable from a formal specification.

The symbolic verifier is deterministic relative to its rule schemas
and diagnostic taxonomy, but this does not make it infallible in an
absolute sense. Expert
annotation supports the validity of these labels, but only on a sample
and cannot exclude all specification or implementation errors. The
fine-tuned detector is likewise trained on verifier-assigned labels and
therefore learns this particular taxonomy.

Our notion of faithfulness is limited to whether a diagnosis is
preserved across pipeline stages. It does not establish that a
rationale reflects the internal causal process by which an LLM formed
its output. The NLI-based rationale evaluator also depends on
expert-crafted hypotheses and may remain sensitive to linguistic
variation. Similarly, the GPT-5.1 feedback judges did not fully
reproduce human ratings, with pedagogical appropriateness showing
especially limited stability. Although two annotators calibrated the
rubric jointly, one annotator scored the remaining sample, limiting the
strength of the human-validation conclusions.

Finally, all generation components use Llama 3.1 8B Instruct and are
explicitly instructed to treat upstream diagnoses and rationales as
authoritative. This design makes error propagation observable, but it
may underestimate the ability of systems permitted to independently
self-correct. We evaluate diagnostic and pedagogical
properties of generated feedback rather than its effect on learning.
A student study is therefore required before concluding that
verifier-grounded feedback improves educational outcomes.
\section*{Ethical Considerations}
The study uses de-identified interaction logs from an undergraduate
propositional-logic tutor. All personal information was removed before
analysis. Although verifier grounding reduces diagnostic uncertainty, generated
feedback may still be incomplete or misleading. The system is intended
to support, rather than replace, instructor judgment and should not be
used as the sole basis for grading or other consequential decisions.

\section*{AI Usage Disclosure}
This work studies and evaluates large language models as research objects. We utilized large language models as assistive tools during manuscript preparation, including formatting guidelines and brainstorming organization, mainly, and paraphrasing on a need basis only. We did not use any AI tools for designing, implementing, or executing this research study. All the claims, analyses, hypotheses, and conclusions are developed, verified, and reviewed by the authors. Moreover, no AI tool was used for generating or labeling data, making judgments about data, or making any scientific claims. All implementations were reviewed, validated, and finalized by the authors. The authors take full responsibility for the correctness, originality, and integrity of the work.


\bibliography{custom}

\appendix
\section{Illustrative Logic Tutor Proof Interaction}
\label{app:DT}

\begin{figure}[H]
    \centering
    \includegraphics[width=\linewidth]{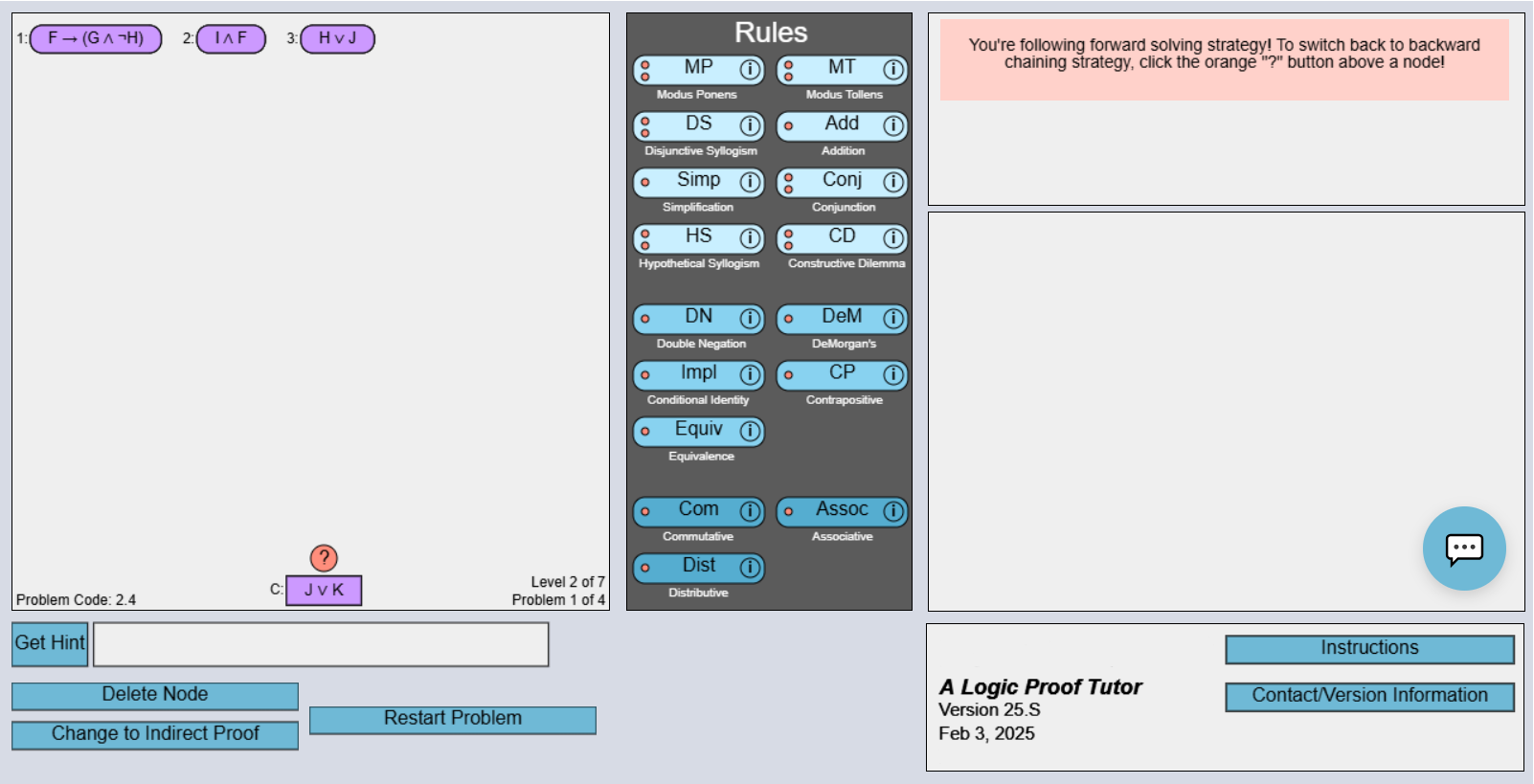}
    \caption{
    \textbf{Initial proof state and goal specification.}
    The student is presented with the premises (top left) and the target conclusion ($J \lor K$) at the bottom. Available inference rules are displayed on the middle. At this stage, no intermediate steps have been derived, and the student must choose a productive forward step toward the goal.
    }
    \label{fig:dt-initial}
\end{figure}

\begin{figure}[H]
    \centering
    \includegraphics[width=\linewidth]{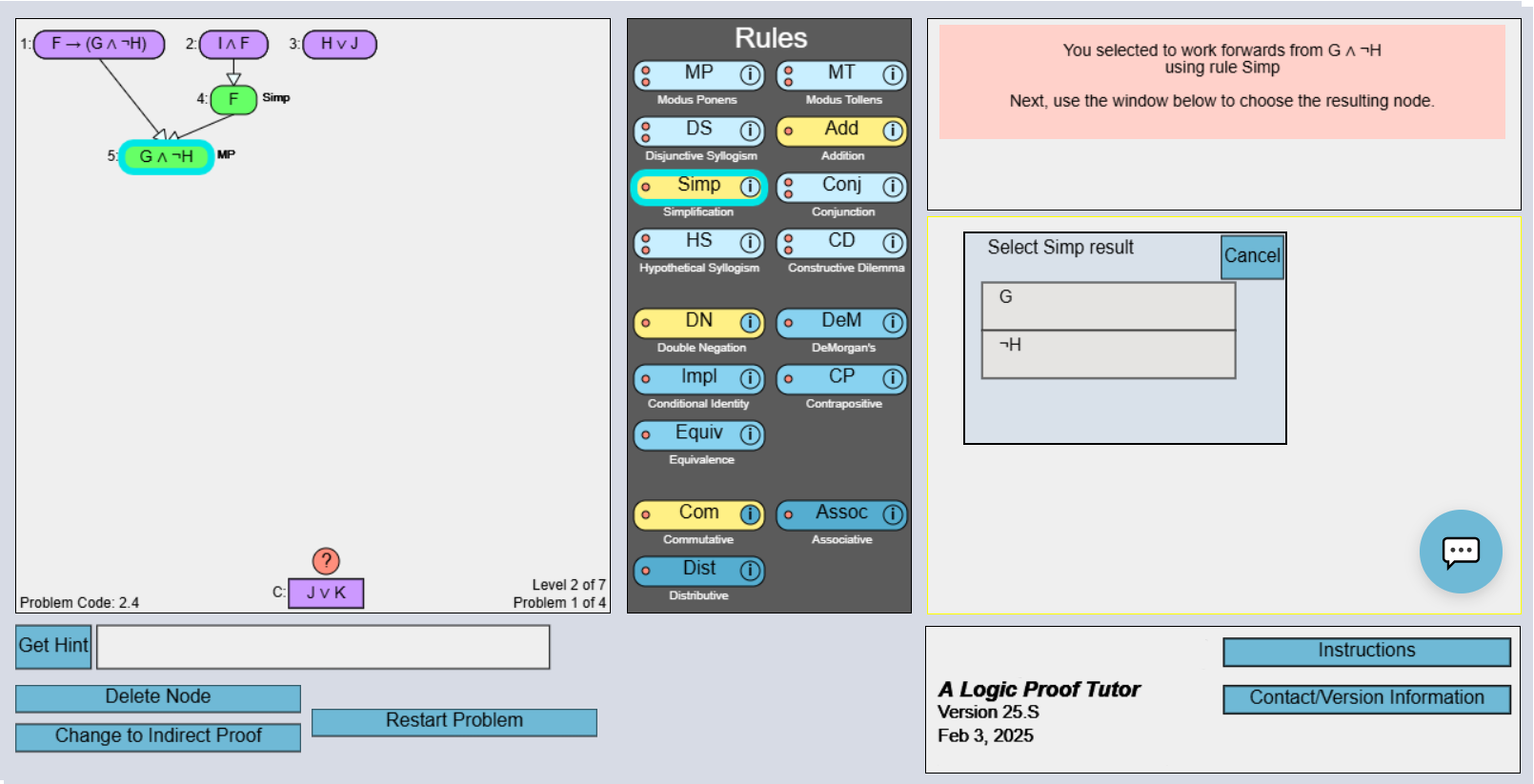}
    \caption{
    \textbf{Rule application with guided simplification.}
    After deriving an intermediate conjunction via Modus Ponens, the student applies the \textit{Simplification} rule. The interface prompts the learner to select the appropriate resulting literal ($G$ or $\lnot H$), illustrating fine-grained, step-level decision making supported by rule constraints.
    }
    \label{fig:dt-simplification}
\end{figure}

\begin{figure}[H]
    \centering
    \includegraphics[width=\linewidth]{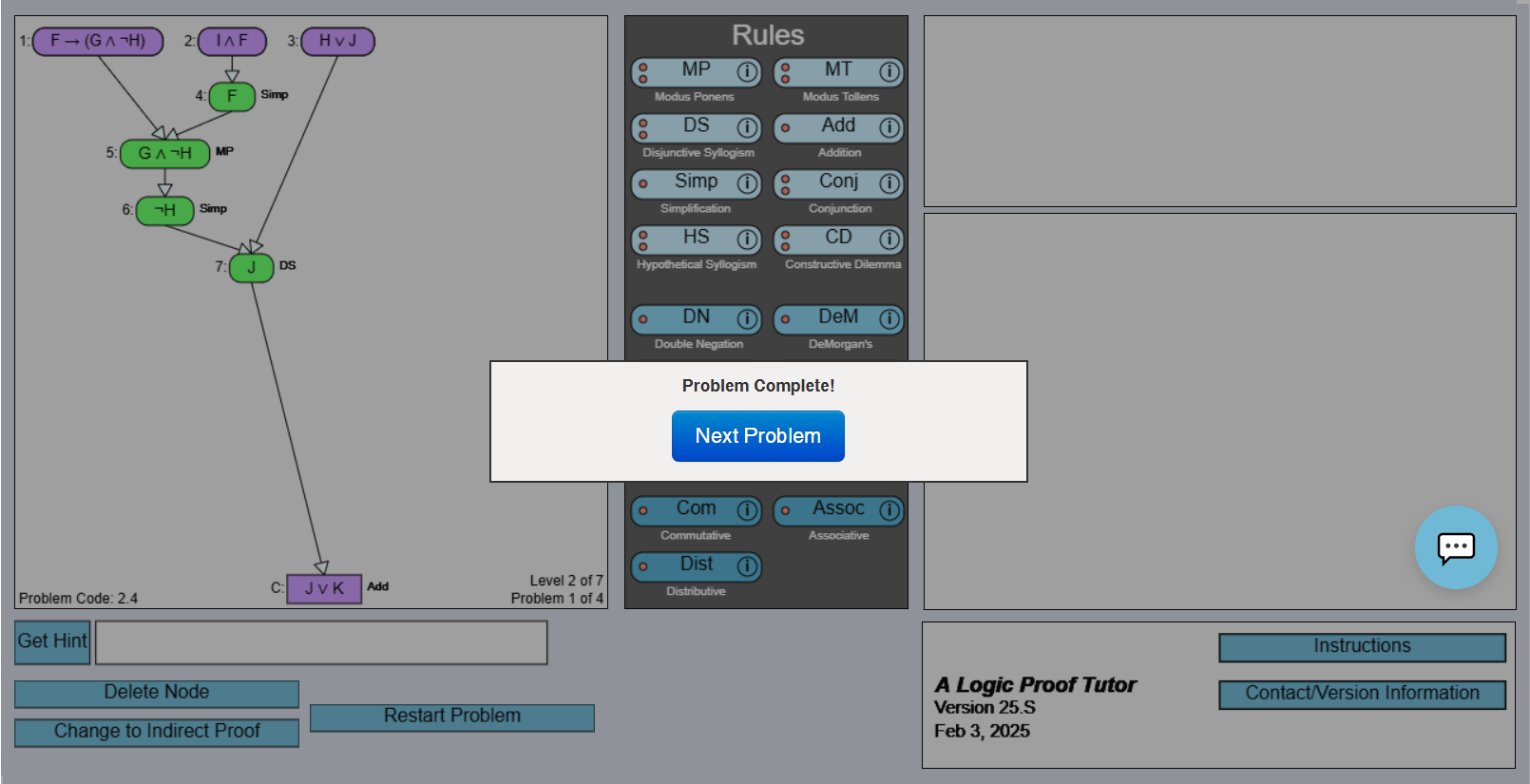}
    \caption{
    \textbf{Successful proof completion.}
    The student derives $J$ via Disjunctive Syllogism and applies the \textit{Addition} rule to reach the target conclusion $J \lor K$. The system confirms completion, reinforcing correct rule sequencing and alignment with the goal state.
    }
    \label{fig:dt-complete}
\end{figure}

Figure~\ref{fig:dt-initial}--\ref{fig:dt-complete} illustrates a representative student interaction in the propositional logic tutor. These screenshots demonstrate forward chaining, rule application, and goal completion within the tutor interface.

\section{Inference Rule List}
\label{appendix:rules}

We employ a fixed set of propositional inference rules used in Logic Tutor for the dataset. The short names were used for consistent response generation and evaluation. The complete list of inference rules, along with short names and derivations are provided below in Table \ref{tab:inference-rules}.

\begin{table}[H]
\centering
\small
\footnotesize
\caption{Propositional inference rules used in this work.}
\label{tab:inference-rules}
\setlength{\tabcolsep}{2pt}
\begin{tabular}{lll}
\toprule
\textbf{Abbrev.} & \textbf{Rule Name} & \textbf{Form} \\
\midrule
MP   & Modus Ponens 
     & $P \rightarrow Q,\; P \Rightarrow Q$ \\

MT   & Modus Tollens 
     & $P \rightarrow Q,\; \neg Q \Rightarrow \neg P$ \\

Conj & Conjunction 
     & $P,\; Q \Rightarrow P \land Q$ \\

Simp & Simplification 
     & $P \land Q \Rightarrow P$ (or $Q$) \\

Add  & Addition 
     & $P \Rightarrow P \lor Q$ \\

DS   & Disjunctive Syllogism 
     & $P \lor Q,\; \neg P \Rightarrow Q$ \\

HS   & Hypothetical Syllogism 
     & $P \rightarrow Q,\; Q \rightarrow R$ \\
     & 
     & $\Rightarrow P \rightarrow R$ \\

Impl & Implication 
     & $P \rightarrow Q \equiv \neg P \lor Q$ \\

DN   & Double Negation 
     & $P \equiv \neg\neg P$ \\

CP   & Contraposition 
     & $P \rightarrow Q \equiv \neg Q \rightarrow \neg P$ \\

Com  & Commutation 
     & $P \lor Q \equiv Q \lor P$ \\

Assoc& Associativity 
     & $(P \lor Q)\lor R$ \\
     & 
     & $\equiv P \lor (Q \lor R)$ \\

Dist & Distribution 
     & $P \land (Q \lor R)$ \\
     & 
     & $\equiv (P \land Q) \lor (P \land R)$ \\

CD   & Constructive Dilemma 
     & $(P\!\rightarrow\!Q),(R\!\rightarrow\!S),P\lor R$ \\
     & 
     & $\Rightarrow Q \lor S$ \\

Equiv& Equivalence 
     & $P \leftrightarrow Q$ \\
     & 
     & $\equiv (P\!\rightarrow\!Q)\land(Q\!\rightarrow\!P)$ \\

\bottomrule
\end{tabular}
\end{table}

\subsection{B1 Fine-Tuning Details}
\label{app:sft_details}

We fine-tune the same base model used in B0 with QLoRA
\citep{dettmers2023qlora} on the 800-instance training set, using LoRA
with rank $r=16$ and $\textit{lora\_alpha}=16$. We use a learning rate
of $2 \times 10^{-4}$ and an effective batch size of 16 (per-device
batch size 2, gradient accumulation steps 8), training for 6 epochs on
a single NVIDIA A30 GPU using the Unsloth framework.

Validation loss was evaluated every half epoch. \texttt{eval\_loss}
decreased substantially through epoch 3.0 (\texttt{eval\_loss} $=
0.01206$), with continued marginal improvement through epoch 5.0
(\texttt{eval\_loss} $= 0.01128$) and diminishing returns thereafter.
We deploy the checkpoint with minimum validation loss, from epoch 5.
Training and evaluation loss curves are shown in
Figure~\ref{app:fig:loss_curves}.

\begin{figure}[h]
\centering
\begin{tikzpicture}
\begin{axis}[
    width=0.85\columnwidth,
    height=6cm,
    xlabel={Epoch},
    ylabel={Loss},
    xmin=0, xmax=6.5,
    ymode=log,
    xtick={0.5,1.0,1.5,2.0,2.5,3.0,3.5,4.0,4.5,5.0,5.5,6.0},
    xticklabel style={font=\small},
    yticklabel style={font=\small},
    legend pos=north east,
    legend style={font=\small},
    grid=major,
    grid style={dashed, gray!30},
]

\addplot[
    color=blue!70,
    mark=circle,
    mark size=2pt,
    thick,
] coordinates {
    (0.5, 0.5908)
    (1.0, 0.03852)
    (1.5, 0.01962)
    (2.0, 0.01463)
    (2.5, 0.01187)
    (3.0, 0.01144)
    (3.5, 0.00975)
    (4.0, 0.00953)
    (4.5, 0.00848)
    (5.0, 0.00843)
    (5.5, 0.00770)
    (6.0, 0.00753)
};
\addlegendentry{Train Loss}

\addplot[
    color=orange!90,
    mark=square,
    mark size=2pt,
    thick,
] coordinates {
    (0.5, 0.05281)
    (1.0, 0.02585)
    (1.5, 0.01662)
    (2.0, 0.01392)
    (2.5, 0.01308)
    (3.0, 0.01206)
    (3.5, 0.01172)
    (4.0, 0.01138)
    (4.5, 0.01169)
    (5.0, 0.01128)
    (5.5, 0.01159)
    (6.0, 0.01142)
};
\addlegendentry{Eval Loss}

\end{axis}
\end{tikzpicture}
\caption{Training and evaluation loss curves for \texttt{llama8b\_error\_classifier\_v5} over 6 epochs.}
\label{app:fig:loss_curves}
\end{figure}

\section{Prompts}
\label{app:prompts}

\subsection{Error Detection Prompt (B0 and B1)}
\label{app:prompt_error_detection}
\begin{small}
\begin{verbatim}
You are an error detector for propositional-logic
proofs.

Given the proof state and student action, return
one diagnosis:
correct, wrong_arity, antecedent_mismatch, or
wrong_rule.

Use the supplied tutor notation, inference-rule
definitions, and arity requirements. Apply the
taxonomy in this order:

1. wrong_arity: the number of selected formulas
   differs from the clicked rule's arity.
2. antecedent_mismatch: the arity is correct and
   the selection fits another approved rule, but
   not the clicked rule.
3. wrong_rule: the arity is correct, but the
   selection fits neither the clicked rule nor
   another approved rule.
4. correct: the selection structurally satisfies
   the clicked rule.

Return only the diagnosis in the required JSON
format.
\end{verbatim}
\end{small}

\subsection{Diagnostic Rationale Agent Prompt}
\label{app:prompt_rationale_agent}

\begin{small}
\begin{verbatim}
You are the diagnostic-rationale component of a
propositional-logic tutor.

You receive a student action and one supplied
diagnosis. Treat the diagnosis as authoritative:
explain it rather than independently reclassifying
the action.

For wrong_arity, explain the mismatch in the
number of selected formulas. For
antecedent_mismatch, explain that the selection
does not fit the clicked rule but matches another
approved rule pattern. For wrong_rule, explain
that the selection is incompatible with the
clicked rule without claiming that another rule
fits. For correct, acknowledge that the selection
satisfies the clicked rule.

Do not name an alternative or corrective rule,
provide a derived formula, or reveal the next proof
step. Produce one or two factual sentences in the
required JSON format.
\end{verbatim}
\end{small}

\subsection{Student Feedback Agent Prompt}
\label{app:prompt_feedback_agent}

\begin{small}
\begin{verbatim}
You are the feedback component of a supportive,
Socratic propositional-logic tutor.

You receive the student
action, and an authoritative rationale. Translate
the rationale into two or three sentences of
student-facing guidance.

Treat the rationale as the sole source of
diagnostic content. Do not independently diagnose
the action, alter the rationale's diagnosis, or
introduce another issue. Use the proof context only
to make the wording concrete.

Provide guidance at the requested hint depth. Do
not name an alternative or corrective rule, give
the exact derived formula, or reveal a complete
next step. Return only the feedback text in the
required JSON format.
\end{verbatim}
\end{small}

\subsection{Feedback Quality Judge Prompt}
\label{app:prompt_feedback_quality_judge}

\begin{small}
\begin{verbatim}
You are evaluating student-facing feedback for a
propositional-logic tutor.

You receive the proof context, student action,
requested hint depth, rationale, and generated
feedback. Treat the rationale as authoritative and
do not independently diagnose the action.

Score each dimension from 0 to 2:

Rationale faithfulness:
2 = clearly preserves the rationale's diagnosis;
1 = compatible but partial or vague;
0 = contradicts or changes the diagnosis.

Non-revelation:
2 = reveals no corrective rule, exact formula, or
    complete next step;
1 = gives a strong partial clue;
0 = explicitly reveals one of these.

Pedagogical appropriateness:
2 = concrete, relevant, and appropriate in depth;
1 = relevant but generic, unclear, or weakly
    actionable;
0 = irrelevant, misleading, or unhelpful.

Return separate scores and brief evidence in the
required JSON format. Do not produce an overall
score.
\end{verbatim}
\end{small}

\subsection{Diagnostic Accuracy Judge Prompt}
\label{app:prompt_diagnostic_accuracy_judge}

\begin{small}
\begin{verbatim}
You are evaluating whether student-facing feedback
communicates the supplied reference diagnosis.

You receive the proof context, student action,
feedback, and verifier-established reference
diagnosis. Do not use a generated rationale or
replace the supplied diagnosis with your own.

Score diagnostic correctness from 0 to 2:

2 = communicates a concrete property specific to
    the reference diagnosis;
1 = compatible with the diagnosis but generic,
    vague, or ambiguous;
0 = contradicts the diagnosis, targets another
    class, or treats an erroneous action as correct.

For wrong_arity, the feedback must address the
number of selected formulas. For
antecedent_mismatch, it must indicate that the
selection does not fit the clicked rule and
corresponds to another rule pattern. For
wrong_rule, it must indicate incompatibility with
the clicked rule without claiming that another
approved rule fits. For correct, it must
acknowledge that the action is valid.

Ignore tone, style, hint depth, and answer
revelation. Return the score and brief evidence in
the required JSON format.
\end{verbatim}
\end{small}

\noindent\fbox{%
\begin{minipage}{\dimexpr\linewidth-2\fboxsep-2\fboxrule\relax}
\textbf{Pipeline:} Proposed symbolic verifier \\
\textbf{Problem ID:} 3.5 \\
\textbf{Rule clicked:} DS \\
\textbf{Formula selected:} $\neg C \rightarrow F$ \\
\textbf{Ground-truth label:} \texttt{wrong\_arity} \\
\textbf{Predicted label:} \texttt{wrong\_arity} \\
\textbf{NLI rationale class:} \texttt{wrong\_arity} \\
\textbf{Hint rationale:}
\textcolor{red}{\textit{The number of formulas selected does not match
the number required by the clicked rule, DS.}} \\
\textbf{Feedback:}
\textcolor{red}{\textit{What is the typical number of formulas required
to apply the rule DS, and how does the number of formulas you selected
compare to that?}} \\
\textbf{Interpretation:} Correct: DS requires two formulas, but only one
was selected.
\end{minipage}%
}

\vspace{1em}

\noindent\fbox{%
\begin{minipage}{\dimexpr\linewidth-2\fboxsep-2\fboxrule\relax}
\textbf{Pipeline:} B1 (SFT) \\
\textbf{Problem ID:} 4.6 \\
\textbf{Rule clicked:} DeM \\
\textbf{Formula selected:} $\neg Y \rightarrow X$ \\
\textbf{Ground-truth label:}
\texttt{antecedent\_mismatch} \\
\textbf{Predicted label:}
\texttt{antecedent\_mismatch} \\
\textbf{NLI rationale class:}
\texttt{antecedent\_mismatch} \\
\textbf{Hint rationale:}
\textcolor{red}{\textit{The complete selection does not fit the clicked
rule, but does fit the structural pattern of a different approved
confusion rule.}} \\
\textbf{Feedback:}
\textcolor{red}{\textit{Consider how the selected formula fits the
structural pattern of a different approved rule, and how that might
affect the proof.}} \\
\textbf{Interpretation:} Correct: the formula fits another rule pattern
rather than DeM.
\end{minipage}%
}

\vspace{1em}

\noindent\fbox{%
\begin{minipage}{\dimexpr\linewidth-2\fboxsep-2\fboxrule\relax}
\textbf{Pipeline:} B1 (SFT) \\
\textbf{Problem ID:} 5.5 \\
\textbf{Rule clicked:} Impl \\
\textbf{Formula selected:} $\neg(\neg G \land B)$ \\
\textbf{Ground-truth label:}
\texttt{antecedent\_mismatch} \\
\textbf{Predicted label:} \texttt{correct} \\
\textbf{NLI rationale class:} \texttt{correct} \\
\textbf{Hint rationale:}
\textcolor{red}{\textit{The selected formula matches the structural
requirement of the clicked rule, Impl.}} \\
\textbf{Feedback:}
\textcolor{red}{\textit{The formula you've selected fits the clicked
rule, Impl.}} \\
\textbf{Interpretation:} Incorrect: B1 accepts a negated conjunction as
a valid implication.
\end{minipage}%
}

\vspace{1em}

\noindent\fbox{%
\begin{minipage}{\dimexpr\linewidth-2\fboxsep-2\fboxrule\relax}
\textbf{Pipeline:} B0 (zero-shot) \\
\textbf{Problem ID:} 2.8 \\
\textbf{Rule clicked:} Simp \\
\textbf{Formula selected:} $\neg(A \rightarrow \neg C)$ \\
\textbf{Ground-truth label:} \texttt{wrong\_rule} \\
\textbf{Predicted label:} \texttt{wrong\_rule} \\
\textbf{NLI rationale class:} \texttt{wrong\_rule} \\
\textbf{Hint rationale:}
\textcolor{red}{\textit{The complete selection is incompatible with the
clicked rule Simp.}} \\
\textbf{Feedback:}
\textcolor{red}{\textit{Consider why the formula
$\neg(A \rightarrow \neg C)$ does not fit the clicked rule Simp.}} \\
\textbf{Interpretation:} Correct: Simp requires a conjunction, not a
negated conditional.
\end{minipage}%
}

\vspace{1em}

\noindent\fbox{%
\begin{minipage}{\dimexpr\linewidth-2\fboxsep-2\fboxrule\relax}
\textbf{Pipeline:} B0 (zero-shot) \\
\textbf{Problem ID:} 5.4 \\
\textbf{Rule clicked:} Simp \\
\textbf{Formula selected:} $B \rightarrow A$ \\
\textbf{Ground-truth label:}
\texttt{antecedent\_mismatch} \\
\textbf{Predicted label:} \texttt{wrong\_arity} \\
\textbf{NLI rationale class:} \texttt{wrong\_arity} \\
\textbf{Hint rationale:}
\textcolor{red}{\textit{The clicked rule Simp requires one formula, but
the student selected two formulas: $B \rightarrow A$.}} \\
\textbf{Feedback:}
\textcolor{red}{\textit{How many formulas does the clicked rule Simp
require, and how many did the student select?}} \\
\textbf{Interpretation:} Incorrect: B0 reports an arity error although
exactly one formula was selected.
\end{minipage}%
}

\subsection{Matched vs. Mismatched Significance Tests (RQ3)}
\label{app:rq3-stats}

Table~\ref{tab:mannwhitney} reports the Mann-Whitney $U$ test comparing
matched and mismatched instances within each pipeline, for each of the
three dimensions in Table~\ref{tab:combined-quality}.

\begin{table}[H]
  \centering
  \small
  \caption{Mann-Whitney $U$ test comparing matched vs.\ mismatched
  instances within B1 and B0, by dimension.}
  \label{tab:mannwhitney}
  \begin{tabular}{llrr}
    \toprule
    \textbf{System} & \textbf{Dimension} & \textbf{$p$} & \textbf{$r$} \\
    \midrule
    B1 & Faithfulness         & $1.05\times10^{-5}$  & 0.204 \\
    B1 & Pedagogy             & $3.36\times10^{-11}$ & 0.288 \\
    B1 & Diagnostic Correctness & $8.01\times10^{-45}$ & 0.676 \\
    \midrule
    B0 & Faithfulness         & $0.178$ & 0.065 \\
    B0 & Pedagogy             & $0.201$ & 0.027 \\
    B0 & Diagnostic Correctness & $9.78\times10^{-36}$ & 0.668 \\
    \bottomrule
  \end{tabular}
\end{table}

\subsection{NLI Hypothesis Templates}
\label{app:nli-templates}

Table~\ref{tab:nli-templates} lists the hypotheses used for the
NLI-based rationale evaluation. We use five manually written
hypotheses for each diagnostic class. The hypotheses paraphrase the
class definitions to reduce dependence on a single wording, while the
same number of hypotheses is used for every class. For each rationale,
the rationale is supplied as the NLI premise and each template as the
hypothesis. Class scores are computed using the maximum entailment
probability, as defined in Section~\ref{sec:rationale-faithfulness}.

\begin{table*}[t]
\centering
\small
\begin{tabular}{p{0.18\textwidth} p{0.76\textwidth}}
\toprule
\textbf{Class} & \textbf{Hypotheses} \\
\midrule

\texttt{wrong\_arity}
&
(1) The student selected the wrong number of formulas for the clicked
rule.
\newline
(2) The rule needs a different number of prior lines than were
selected.
\newline
(3) Too few or too many formulas were selected for this rule.
\newline
(4) The number of selected formulas does not match what the rule
requires.
\newline
(5) The student selected an incorrect count of formulas.
\\[3pt]

\texttt{antecedent\_mismatch}
&
(1) The student's selected formulas do not satisfy the structural
pattern of the clicked rule.
\newline
(2) The selected formulas do not satisfy the structural pattern of the
clicked rule.
\newline
(3) The formulas do not match the antecedent requirements of the
clicked rule.
\newline
(4) The arity is correct but the formulas do not structurally match the
clicked rule.
\newline
(5) The clicked rule cannot be applied because the formula types do not
match its pattern.
\\[3pt]

\texttt{wrong\_rule}
&
(1) The student chose an inference rule that does not apply to the
selected formulas.
\newline
(2) The clicked rule cannot produce a valid derivation from these
formulas.
\newline
(3) A different rule should have been chosen for these selected
formulas.
\newline
(4) The student applied the wrong inference rule to otherwise valid
formula selections.
\newline
(5) The rule clicked does not match the logical form of the selected
formulas.
\\[3pt]

\texttt{correct}
&
(1) The student's attempted proof step is correct.
\newline
(2) The selected formulas validly support applying the clicked rule.
\newline
(3) The clicked rule is correctly applied to the selected formulas.
\newline
(4) The student's selected formulas satisfy the requirements of the
clicked rule.
\newline
(5) The attempted inference is valid and contains no structural error.
\\
\bottomrule
\end{tabular}
\caption{Hypotheses used for the NLI-based assignment of generated
rationales to the four diagnostic classes. Each generated rationale is
used as the NLI premise, and each sentence listed above is evaluated as
a hypothesis.}
\label{tab:nli-templates}
\end{table*}

\section{Annotation Protocol}
\label{app:annotation}

\paragraph{Error-taxonomy validation.}
Two annotators independently labeled a stratified sample of 100
student actions while blind to the verifier's outputs. Using the
definitions of \texttt{correct}, \texttt{wrong\_arity},
\texttt{antecedent\_mismatch}, and \texttt{wrong\_rule}, they assigned
one diagnosis to each action. Inter-annotator agreement exceeded
$\kappa=.80$; agreement with the verifier was 95\% and 92\% for the
two annotators, respectively.

\paragraph{Rationale validation.}
One annotator reviewed a stratified sample of 200 rationales
(50 per assigned diagnosis) and judged whether each clearly preserved
the diagnosis it was generated to explain.

\begin{figure*}[p]
    \centering
    \includegraphics[
        width=\textwidth,
        height=0.88\textheight,
        keepaspectratio
    ]{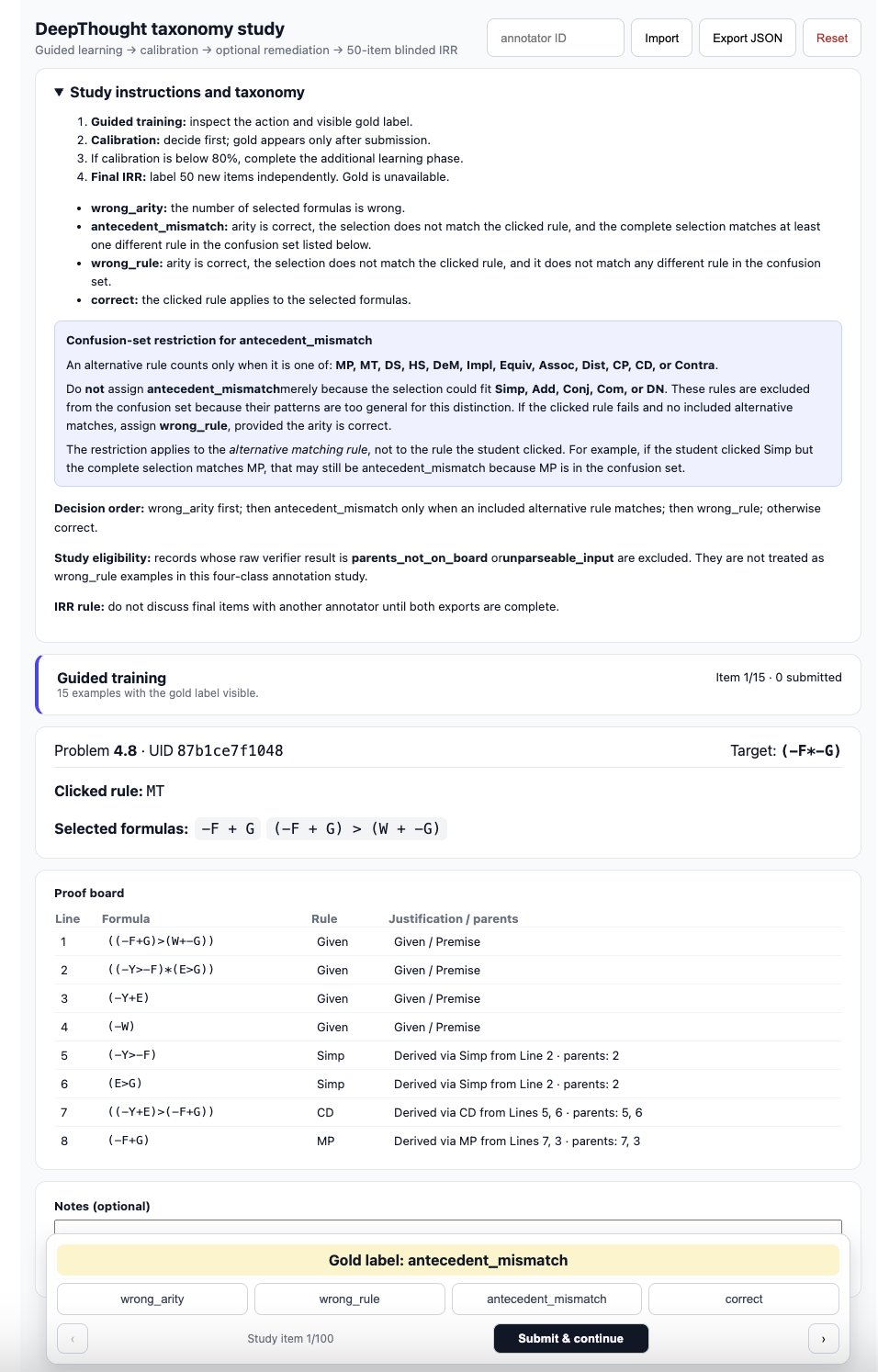}
    \caption{Interface used for error-taxonomy annotation.}
    \label{fig:annotation-error}
\end{figure*}

\begin{figure*}[p]
    \centering
    \includegraphics[
        width=\textwidth,
        height=0.88\textheight,
        keepaspectratio
    ]{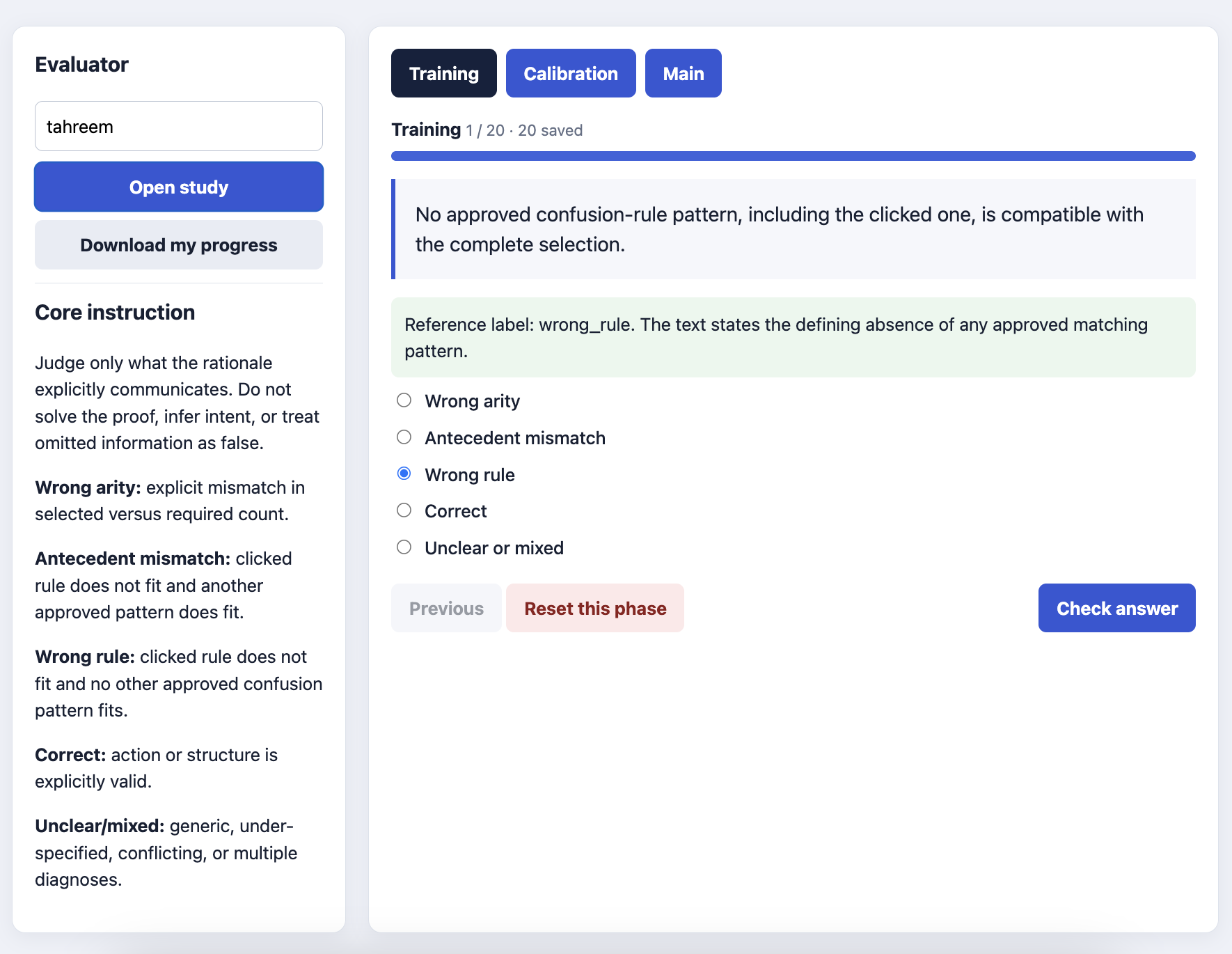}
    \caption{Interface used for rationale-faithfulness annotation.}
    \label{fig:annotation-rationale}
\end{figure*}

\begin{figure*}[p]
    \centering
    \includegraphics[
        width=\textwidth,
        height=0.88\textheight,
        keepaspectratio
    ]{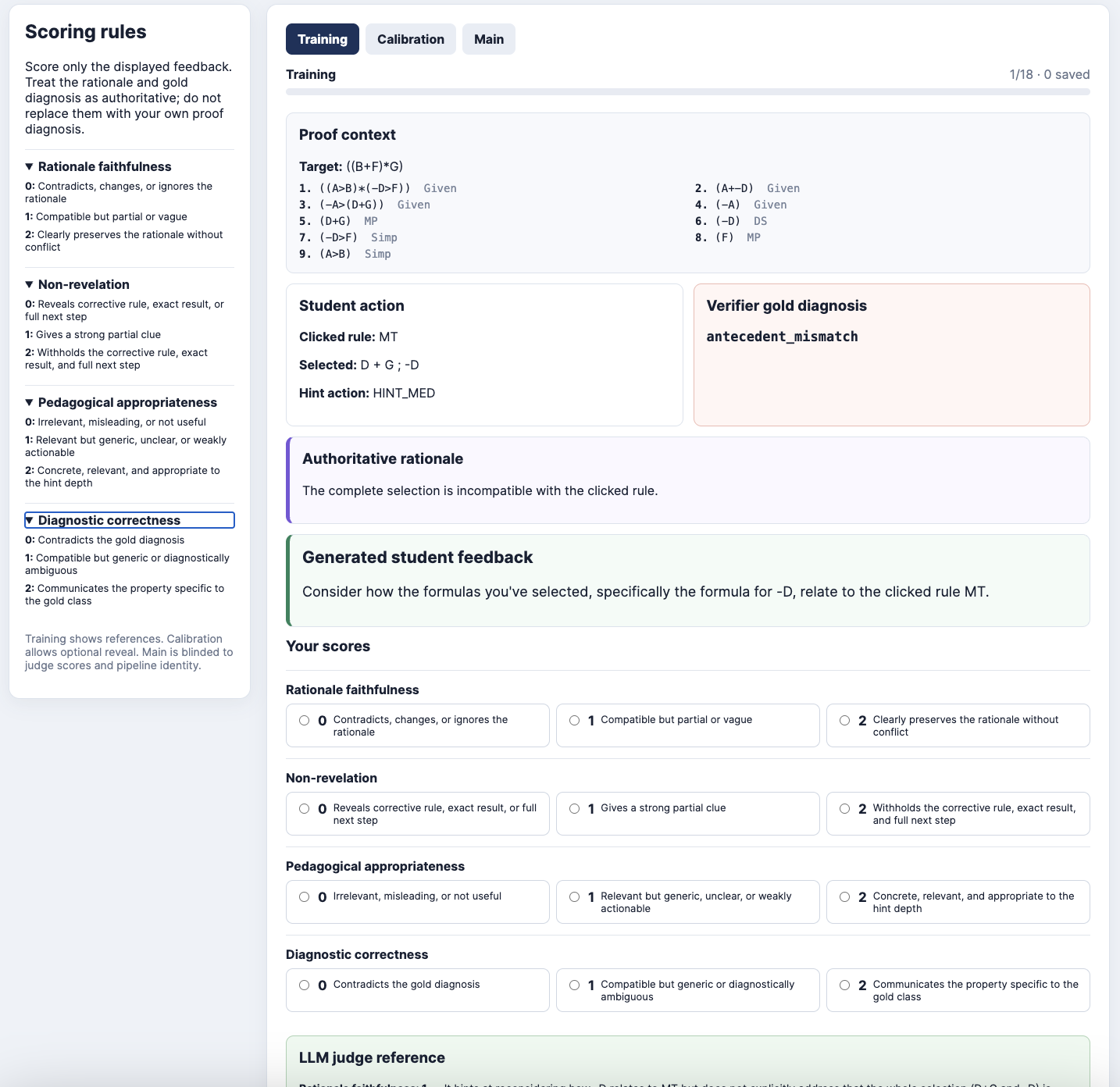}
    \caption{Interface used for feedback annotation.}
    \label{fig:annotation-feedback}
\end{figure*}

\paragraph{Feedback validation.}
We sampled 150 feedback messages across pipelines and diagnosis
classes. Two annotators independently scored 50 calibration messages
using the same rubric as the automatic judges while blind to their
outputs. After training and calibration, they reached $\kappa=.89$ agreement.
One annotator then scored the remaining 100 messages. Human ratings
were compared separately with the quality- and diagnostic-judge
scores. The instructions and the applications set up for each of the annotation is as follows: Figure~\ref{fig:annotation-error} shows the error-taxonomy annotation
interface, Figure~\ref{fig:annotation-rationale} the rationale
annotation interface, and Figure~\ref{fig:annotation-feedback} the
feedback annotation interface.

\clearpage

\clearpage

\end{document}